\documentclass[10pt,twocolumn,letterpaper]{article}

 \usepackage{cvpr}              % To produce the CAMERA-READY version
\usepackage[dvipsnames]{xcolor}

\definecolor{cvprblue}{rgb}{0.21,0.49,0.74}
\usepackage[pagebackref,breaklinks,colorlinks,citecolor=cvprblue]{hyperref}

\def\paperID{13552} % *** Enter the Paper ID here
\def\confName{CVPR}
\def\confYear{2024}

\title{Multi-label Learning from Privacy-Label}

\author{
  Zhongnian Li \qquad Haotian Ren \qquad Tongfeng Sun\thanks{Corresponding author.}  \qquad Zhichen Li \\
  School of Computer Science and Technology, China University of Mining and Technology\\
  Xuzhou, Jiangsu, China \\
  {\tt\small \{zhongnianli, ts22170042a31, suntf, ts22170089p31\}@cumt.edu.cn}
}

\usepackage{multirow} 
\usepackage{setspace}
\usepackage{booktabs,multirow}
\usepackage{url}
\usepackage{tabu}                     % 表格插入
\usepackage{multirow}                 % 一般用以设计表格，将所行合并
\usepackage{multicol}                 % 合并多列
\usepackage{multirow}                % 合并多行
\usepackage{float}                    % 图片浮动
\usepackage{makecell}                 % 三线表-竖线
\usepackage{booktabs}                 % 三线表-短细横线
\usepackage[ruled,linesnumbered]{algorithm2e}
\usepackage{colortbl}
\usepackage{caption}

\let\oldtwocolumn\twocolumn
\renewcommand\twocolumn[1][]{%
    \oldtwocolumn[{#1}{
    \begin{center}
           \includegraphics[width=0.9\textwidth]{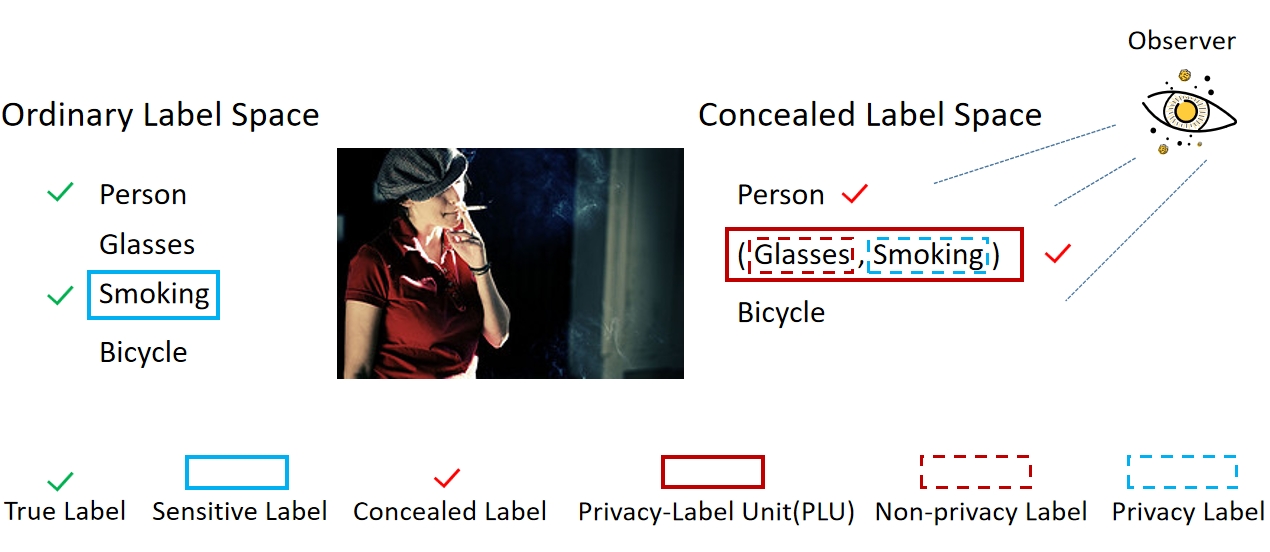}
           \captionof{figure}{Illustrates an example of labeling sensitive labels through PLU in NUS-WIDE dataset when the annotator is being observed.  The image has true labels including ``Person" and ``Smoking", where ``Smoking" is sensitive label. To ensure the sensitive label remains undisclosed in the label set, Privacy-Label Unit (PLU) combines ``Smoking" and a randomly sampled label ``Glasses" into a single unit, which only ``Glasses" exist in the label set. By using PLU, only ``Person", ``Bicycle", ``Glasses and Smoking" will be observed by observer. In this example, the observer cannot determine whether the true label is ``glasses," ``smoking," or both. In the unit, ``Glasses" is named non-privacy label and ``Smoking" is named privacy-label. The label of PLU is positive, due to the true label of ``Smoking" is positive. On the contrary, if both the true label of ``Glasses" and ``Smoking" are negative, then the label of PLU is regarded as negative.}
           \label{fig:one}
        \end{center}
    }]
}

\begin{document}
\maketitle

\begin{abstract}
Multi-abel Learning (MLL) often involves the assignment of multiple relevant labels to each instance, which can lead to the leakage of sensitive information (such as smoking, diseases, etc.) about the instances. However, existing MLL suffer from failures in protection for sensitive information. In this paper, we propose a novel setting named Multi-Label Learning from Privacy-Label (MLLPL), which Concealing Labels via Privacy-Label Unit (CLPLU). Specifically, during the labeling phase, each privacy-label is randomly combined with a non-privacy label to form a Privacy-Label Unit (PLU). If any label within a PLU is positive, the unit is labeled as positive; otherwise, it is labeled negative, as shown in Figure \ref{fig:one}. PLU ensures that only non-privacy labels are appear in the label set, while the privacy-labels remain concealed. Moreover, we further propose a Privacy-Label Unit Loss (PLUL) to learn the optimal classifier by minimizing the empirical risk of PLU. Experimental results on multiple benchmark datasets demonstrate the effectiveness and superiority of the proposed method.
\end{abstract}
    
\section{Introduction}
\label{sec:intro}

Multi-label Learning (MLL) aims to train a classifier that can classify instances with multiple labels. Unlike multi-class learing, each data instance in MLL can be associated with multiple labels, providing a more nuanced representation that captures the diversity and complexity of the data.
In recent years, the scope of MLL applications has expanded significantly, demonstrating its versatility across various domains. In natural language processing, MLL has proven valuable for tasks involving document categorization, sentiment analysis, and topic labeling \cite{MLL26,MLL27}. In image recognition, especially in the context of scene understanding and object recognition, MLL enhances the ability to recognize complex scenes or images with diverse content \cite{MLL30,MLL31}. Additionally, in bioinformatics, MLL contributes to the annotation of biological sequences, enabling more comprehensive analysis and interpretation of genomic data \cite{MLL28}. In the field of medical image analysis, MLL facilitates the diagnosis of diseases by allowing the simultaneous consideration of multiple relevant features in medical images \cite{MLL29}.

In traditiona multi-label learning, it is costly and nay impossible in some real-world scenarios to acquire massive amount of high-quality labeled data. To tackle this challenge, various weakly supervised framework has been proposed recently, aiming to investigate the feasibility of training models with incomplete labels. Such as partial multi-label (PLM) \cite{2020PartialML19, PML20}, which address the scenario where only partial labels are available. Multi-label learning with missing labels (MLML) \cite{MLML21, MLML22}, which specifically focuses on scenarios where some labels are missing, offering a solution to the challenges posed by incomplete labeling in training datasets. Additionally, single-positive multi-label learning (SPML) is designed to train models with only positive labels provided for each training example, eliminating the need for additional negative or positive labels \cite{17,18}. Moreover, multi-labeled complementary label learning (MLCLL) deal with instances associated with a set of complementary labels, capturing both presence and absence information of certain attributes \cite{MLCLL23}.

Unfortunately, prevailing weakly supervised learning methods, including PLM, MLML, SPML, and MLCLL, tend to overlook a critical challenge: the protection of sensitive labels. As illustrated in Figure \ref{fig:one}, labels like `smoking' are intricately connected to personal choices involving health, lifestyle, and social circles. Consequently, the label annotation phase presents a significant challenge, as it carries the inherent risk of inadvertently revealing private information about an individual's life, encompassing details about health habits and behavioral choices. This inadvertent leakage poses a serious threat to individual privacy, highlighting the need for robust measures to protect sensitive information. In practical scenarios, the information related to smoking may be inadvertently leaked during the label annotation phase when it is observed. 

In this paper, we explore the setting of Multi-Label Learning from Privacy-Labels (MLLPL) problem, which aims to prevent privacy labels from appearing in the label set. As shown in Figure \ref{fig:one}, we introduce the Privacy-Label Unit (PLU) to conceal privacy-labels by combining a privacy-label (Smoking) with a randomly sampled non-privacy label (Glasses). 
The label of PLU is treated as negative when both privacy label and non-privacy label in the unit are negative, otherwise it is treated as positive. Attributed to the characteristics of PLU, it ensures that privacy labels will not be disclosed in the data. For privacy-label unit, we further introduce privacy-label unit loss by optimizing the setting of PLU label to recover the hidden labels. The contributions of this paper can be summarized as follows:
\begin{itemize}
    \item We propose a privacy-label setting method for multi-label learning with privacy labels, introducing the concept of the ``Privacy-Label Unit (PLU)''. PLU is designed to protect the privacy of sensitive information during the label collection stage.
    \item We propose a specific loss functions to offer efficient algorithms for learning with privacy-labels.
    \item Extensive experiments on 11 benchmark datasets show that the classifiers learned through our method can effectively recognize instances with concealment of labels.
\end{itemize}

\section{Related Work}
\label{sec:formatting}
\subsection{Multi-label Learning}
Differing from multi-class classification, the output space in multi-label learning grows exponentially with the number of labels. To address the challenges posed by this expanding output space, many algorithms focus on uncovering correlations between labels. Based on the degree of correlation mining \cite{1}, multi-label learning approachs can be categorized into three distinct groups. 

The first-order approach ignores the correlation with other labels, it decomposes the multi-class problem into multiple independent binary classification problems. \cite{2} introduces a classic  first-order approach, which nemad ML-KNN. That is a lazy learning approach derived from K-nearest neighbor, which predicts label sets for unseen instances based on statistical information from their K nearest neighbors. \cite{3} introduces multi-label learning with Label specific Features (LIFT), which enhances discrimination among different class labels by constructing label-specific features through clustering analysis on positive and negative instances.. This method is simple and efficient. 

The second-order approach consider the correlation between label pairs \cite{4,5,6}. \cite{6} proposes a Classifier Chain Model (CC), which passes label information between classifiers and thus overcoming the label independence problem. This approach extends beyond first-order methods by accounting for dependencies between labels in pairs, contributing to more accurate predictions in multi-label scenarios. 

The high-order approach takes into account the influence of all other labels in the label set for each label \cite{7,8,9}. Compared to the previous two methods, the high-order approach yields better results. Recently, many methods employ Graph Convolutional Network (GCN) \cite{12} to capture the structure information between multiple labels \cite{11}. ML-GCN \cite{10} establishes a co-occurrence matrix with labels as nodes to capture co-occurrence correlations among labels.

\subsection{Weakly Supervised Multi-label Learning}
In practice, traditional multi-label learning demands precise annotations of all relevant labels for each instance, which is expensive or even impossible, particularly when dealing with instance privacy. To reduce the cost of the label annotation phase, numerous weakly supervised multi-label learning frameworks have been studied, including, MLML,PML,SPML,MLCLL, etc.

PML addresses problems with complex semantic labels \cite{PML32}. For example, it is usually hard to decide whether an iris is an Iris setosa or an Iris versicolor. PML annotates all possible labels as positive labels, where only a subset of them are truly relevant labels, and the remaining labels are false negatives \cite{14zongshu,15}. 
For example, \cite{PML33} introduces a regularization training framework, conducting supervised learning on non-candidate labels and applying consistency regularization on candidate labels. In \cite{PML34}, PML problem is addressed by initially employing label propagation algorithm to disambiguate the candidate, and then training the model on processed dataset. By comparison, not complete positive labels are requested in MLML, a subset of negative labels are false negative, i.e., missing labels \cite{16}. \cite{MLML35} solves MLML problem via dependency graphs, which treats each label of each instance as a node. Nevertheless, MLML will not work if one label is completely unobserved \cite{15}.

In SPML, annotators are requested to offer only one positive label for each training example, without the need for additional negative or positive labels \cite{17}. Prior methods trains the model by assuming unobserved labels as negatives \cite{18} and employed regularization to control the count of anticipated positive labels. Recently, many studies have addressed SPML problems by recovering potential positive labels \cite{18,17}.
For example, \cite{SPML25} proposed a Label-Aware global Consistency (LAC) regularization to regain information about potential positive labels. \cite{17} employed label enhancement to obtain soft labels and trained classification models by minimizing the empirical risk estimate. 

In MLCLL, there is a set of complementary labels associated with each instance. Complementary labels refer to the absence or negation of certain attributes or characteristics present in the main labels. \cite{MLCLL23} proposed an unbiased risk estimator with an estimation error bound via minimizing learned risk to learn a multi-label classifier from complementary labeled data.

Unfortunately, current state-of-the-art weakly supervised multi-label learning methods cannot well address the challenge of label privacy leakage. In MLLPL problem, the loss of privacy-label (e.g., smoking) will lead to a significant decrease in the model's generalization performance, which is different from MLML that lacks certain accurate labels. In this case, existing MLML approaches may not be suitable for MLLPL, as they are built on the assumption that each labels could potentially appear. Similarly, SPML will not work if a certain label is entirely missing. MLCLL trains the classifier by annotating negative labels. Although it conceals the sensitive information of positive labels during the data annotation stage, it fails to provide privacy protection for negative labels.

\section{The Proposed Method}
\subsection{Problem Setup}
\textbf{Multi-label learning.} In the standard MLL, each instance requires assignment of multiple relevant labels. Let $\mathcal Z=\left\{0,1\right\}^{L}$ be the target label space with L classes labels and $\mathcal X\in\mathbb{R}^{d}$ be the feature space with d dimensions. Given fully-supervised dataset $D=\left\{\left ( x_{i},z_{i} \right )\right\}_{i=1}^{n}$ which indicates that each $x_{i}\in\mathcal{X}$ is associated with a vector of label $z_{i}^{j} \in \mathcal{Z}$, where an entry $z_{i}^{j}=1$ if the \textit{j}-th class is relevant to $x_{i}$ and $z_{i}^{j}=0$ if the \textit{j}-th label is not relevant. The aim for MLL is to induce a classifier $f:\mathcal{X} \to \left [ 0,1 \right ]^{L}$ by minimizing the following classification risk:
\begin{equation}\label{eq1}
  R(f)=\mathbb{E}_{p(x,z)}\big[\mathcal L(f(x),z)\big],
\end{equation}
where $\mathcal{L}$ is the loss function.

Additionally, we employ binary cross-entropy (BCE) loss as the loss function $\mathcal{L}$. BCE loss is the predominant choice in traditional multi-label classification, it essentially treats each label as an individual binary classification task and solves the multi-label problem by independently applying the BCE loss to each label. Given a batch of MLL examples $\left\{\left ( x_{i},z_{i} \right )\right\}_{i=1}^{b}$, $\mathcal{L}$ can be expressed as follows:
\begin{equation}\label{eq2}
    \begin{split}
     \mathcal {L}=-\frac{1}{b}\sum_{i=1}^{b}\sum_{j=1}^{L}(1-z_{i}^{j})log(1-p_{i}^{j})+z_{i}^{j}log(p_{i}^{j}),
    \end{split}
\end{equation}
where $p_{i}^{j}$ denotes $p(z_{i}^{j}|x_{i})$, respectively. Clearly, $p(z_{i}^{j}|x_{i})$ denotes the conditional distribution of $f(x_{i})$ at $j$-th label. For simplicity, in the sequel we ignore
the subscript $i$ in $p_{i}^{j}$ and $p_{i}^{j}$ to use $p_{j}$ and $p_{j}$.

\textbf{Multi-label learning from privacy label.} For the MLLPL problem studied in this paper, we we split label into two components, $y \in \left\{0,1\right\}^c$ and $\bar{y} \in \left\{0,1,\O \right\}^m$ where $y$ denotes the observed labels, $c$ is the number of the observed labels in label set and $\bar{y}$ is the privacy-label unit labels, $m$ is the number of the labels in PLUs. The MLLPL dataset can be denoted as $\tilde{D}=\left\{\left ( x_{i},y_{i},\bar{y_{i}} \right )\right\}_{i=1}^{n}$. For each MLLPL training example $\left ( x_{i},y_{i},\bar{y_{i}} \right )$, the observed labels vector is denoted by $y_{i}=\left [ y_{i}^{1},y_{i}^{2},...,y_{i}^{c} \right ]^{T} \in \left\{0,1\right\}^c$, where $y_{i}^{j}=1$ if the $j$-th label is revelant to $x_{i}$, otherwise, $y_{i}^{j}=0$. The label vector in PLU are represented as $\bar{y}_{i}=\left [ \bar{y}_{i}^{1},\bar{y}_{i}^{2},..., \bar{y}_{i}^{m}\right ]^{T} \in \left\{0,1,\O \right\}^c$, $\bar{y}_{i}^{p}=\o $ indicates that the $p$-th label is privacy-label. The non-privacy label paired with privacy-label is denoted as $\bar{y}_{i}^{s}$, $\bar{y}_{i}^{s}=$ PLU $ \in \left\{0,1\right\}$, as shown in Figure \ref{fig:one}. If $z_{i}^{s}=z_{i}^{p}=0$, where $z_{i}^{s}$ and $z_{i}^{p}$ is the ordinary labels corresponding to  $\bar{y}_{i}^{s}$ and $\bar{y}_{i}^{p}$, PLU$=\bar{y}_{i}^{s}=0$, otherwise PLU$=\bar{y}_{i}^{s}=1$. Here PLU$=\bar{y}_{i}^{s}$ contains the labeling information common to privacy-label and non-privacy label. 

\subsection{Conceal Label via Privacy-Label Unit}
In this seciton, we provide the privacy-label unit loss for the CLPLU method. Based on the MLLPL setup of the previous section and eq.(\ref{eq1}), $R_{clplu}(f)$ can be expressed as the sum of the fully-supervised classification risk $R_{fu}(f)$ of observed labels and the weakly-supervised classification risk $R_{plu}(f)$ of labels in PLUs:
\begin{equation}\label{eq3}
    R_{clplu}(f)=R_{fu}(f)+R_{plu}(f),
\end{equation} 
where $R_{clplu}(f)$ is the classification risk of our method and $f$ is the multi-label classifier. Given the observed labels $y$ and the privacy-label unit labels $\bar{y}$, fully-supervised classification risk and weakly-supervised classification risk can be defined as

\begin{equation}\label{eq4}
    R_{fu}(f)=\mathbb{E}_{p(x,y)}\big[\mathcal L_{fu}(f(x),y)\big],
\end{equation} 
\begin{equation}\label{eq5}
    R_{plu}(f)=\mathbb{E}_{p(x,\bar{y})}\big[\mathcal L_{plu}(f(x),\bar{y})\big],
\end{equation} 
where $\mathcal{L}_{fu}$ and $\mathcal{L}_{plu}$ denote the loss functions of fully-supervised learning and weakly-supervised learning, separately. We use the sigmoid method for $f($\textbullet$)$ to approximate the value of the probability distribution. In the next subsections, we present the details of $\mathcal{L}_{fu}$ and $\mathcal{L}_{plu}$.

\subsection{Fully-supervised Loss}
For fully supervised classification part, we directly employ traditional BCE loss as its loss function. Then, $\mathcal L_{fu}(f(x),y)$ in eq.(\ref{eq4}) can be replaced by $\mathcal {L}(f(x),y)$. For each CLPLU example $\left ( x_{i},y_{i},\bar{y}_i \right )$:

\begin{equation}\label{eq6}
   % \mathcal{L}_{fu}(f(x),y)= -\sum_{j \in Y}log(1-p_{j}^{+})+log(p_{j}^{-}),
    \mathcal{L}_{fu}(f(x),y)=\sum_{j \in y_{i}}(1-y_{i}^{j})log(1-p_{j})+y_{i}^{j}log(p_{j}).
\end{equation}

Here, $p_{j}$ denotes $p(y_{i}^{j}|x_{i})$, and the same applies to the follows. However, $\mathcal L_{plu}(f(x),\bar{y})$ in eq.(\ref{eq5}) cannot directly employ the conventional BCE loss. The PLU conceals the privacy-labels which implies that the model cannot acquire information related to these labels. To address the issue, we propose three variants of BCE loss to alleviate the negative impact of missing labels.

\subsection{Weakly-supervised Loss}
In this section, we present the specific details of three variants of BCE. We first introduce two intuitive approaches to modify BCE and provide corresponding loss functions, namely AN loss and AP loss. Than, we propose the Privacy-Label Unit Loss(PLUL) and explain its working principles.

\textbf{Assuming negative for PLU.} In real multi-label scenarios, the number of positive labels is often much smaller than the number of negative labels. Under this premise, a straightforward method to mitigate the effect of missing
labels is to treat all labels in the PLU as negative, i.e., $\bar{y}_{i}^{j}=0$ where $\bar{y}_{i}^{j} \in \bar{y}$. For each CLPLU example $\left ( x_{i},y_{i},\bar{y}_i \right )$:

\begin{equation}\label{eq7}
    \mathcal {L}_{plu}^{AN}=-\sum_{j \in \bar{y}_{i}}log(1-p_{j}^{-}),
\end{equation}
where $p_{i}^{-}$ denotes $p(\bar{y}_{i}^{j}=0|x_{i})$ and $j$ deontes the $j$-th label in $\bar{y}$. The AN loss prevents the model from degrading into trivial solutions that offer no meaningful insights. 

However, numerous false negative labels lead to a significant decrease in generalization performance of the model. In the following, we introduce AN loss to address this issue.

\textbf{Assuming positive for PLU.}  We attempt to recover some positive labels. Specifically, for each PLU in concealed label set, if the unit is negative, then let $\bar{y}_{i}^{s}=\bar{y}_{i}^{p}=0$, else if PLU = 1 which indicates that at least one positive label exists in the unit, then let $\bar{y}_{i}^{s}=\bar{y}_{i}^{p}=1$. The loss function can be defined as:
\begin{equation}\label{eq8}
    \mathcal{L}_{plu}^{AP}\!=\!-\!
    \begin{cases}
     \!\sum\limits_{(s,p) \in \bar{y}_{i}}log(1\!-\!p_{s}^{-})\!+\!log(1\!-\!p_{p}^{-}),  &\text{if PLU}=0\\[12pt]
   \!\sum\limits_{(s,p) \in \bar{y}_{i}}log(p_{s}^{+})+log(p_{p}^{+}),  &\text{if PLU}=1\\[12pt]
    \end{cases}
\end{equation}
where $(s,p)$ denotes non-privacy label and privacy-label belonging to the same PLU. $p_{i}^{+}$ denotes $p(\bar{y}_{i}^{j}=1|x_{i})$.

AN loss and AP loss are two of the most intuitive methods to handle unobserved labels. Regrettably, those methods introduces a significant number of false negative labels or false positive labels, which leads to overfitting to incorrect labels, thereby reducing the model's performance and generalization capability.

\textbf{Privacy-Label unit loss.}  To mitigate the negative impact of
false negatives, we further design the PLUL method. Specifically, we first check if PLU equals 0, which indicates that both labels in the unit are negative. In this case, we set $\bar{y}_{i}^{s}=\bar{y}_{i}^{p}=0$. If PLU is equal to 1, it indicates that at least one label in PLU is positive, that leads to three possible scenarios: a) $\bar{y}_{i}^{s}=1 \text{ and }\bar{y}_{i}^{p}=0$; b) $\bar{y}_{i}^{s}=0 \text{ and }\bar{y}_{i}^{p}=1$; c) $\bar{y}_{i}^{s}=1 \text{ and }\bar{y}_{i}^{p}=1$. Therefore, the loss with the lowest risk among the three situations is selected to train the model. PLUL loss can be formulated as:
\begin{equation}\label{eq9}
    \begin{split}
    \mathcal{L}_{PLUL}\!=\!\!-\!
    \begin{cases}
      \!\sum\limits_{(s,p) \in \bar{y}_{i}}\!log(\!1\!-\!p_{s}^{-})\!+\!log(\!1\!-\!p_{p}^{-}), \!&\text{if PLU}\!=\!0\\[10pt]
   \!\sum\limits_{(s,p) \in \bar{y}_{i}}\! \min(G_{-}^{+},G_{+}^{-},G_{+}^{+}), \!&\text{if PLU}  \!=\!1
    \end{cases}
    \end{split}
\end{equation}
where $G_{-}^{+}=log(p_{s}^{+})\!+\!log(1\!-\!p_{p}^{-})$, $G_{+}^{-}=log(p_{s}^{-})\!+\!log(1\!-\!p_{p}^{+})$, $G_{+}^{+}=log(p_{s}^{+})\!+\!log(1\!\!-p_{p}^{+})$ and 
$\min$ denotes as the minimum risk among the three scenarios a), b) and c).

In order to show the details of the proposed method more intuitively, we show the model learning process in Algorithm \ref{alg:algorithm1}.

\begin{algorithm}
\caption{Conceal Label via Privacy-Labe Unit}
% \begin{algorithmic}
\label{alg:algorithm1}
\KwIn{

$\tilde{D}=\left\{\left ( x_{i},y_{i},\bar{y_{i}} \right )\right\}_{i=1}^{n}$ : The CLPLU training data;

$\mathcal{T}$: The number of epoch;

$\mathcal{K}$: The class number of $\tilde{D}$;

$\mathcal{P}$: The label indexes of privacy-label unit;}
\KwOut{
$\Theta$ : model parameter for $f(x;\Theta)$;}
\BlankLine
\emph{
\For{t=\rm 1,...,$\mathcal{T}$}{
        \For{k=\rm 1,...,$\mathcal{K}$}{
            \eIf{k \rm in $\mathcal{P}$}
                {\rm calculate the loss by eq.(\ref{eq9})\;
                    $\mathcal{L}_{PLUL}\leftarrow \mathcal{L}_{PLUL} +  \mathcal{L}_{PLUL}^{k}$\;}
                {\rm calculate the loss by eq.(\ref{eq6})\;
                    $\mathcal{L}_{fu}\leftarrow \mathcal{L}_{fu} +  \mathcal{L}_{fu}^{k}$\;}
                    $\widehat{R}(f)\leftarrow \mathcal{L}_{PLUL}+\mathcal{L}_{fu}$\rm \;
           \rm Update $\Theta$ via gradient descent\;}
}
}
\end{algorithm}

\section{Experiments}
\subsection{Experiments settings}

\textbf{Dataset.} We perform experiments on eleven widely-used MLL datasets to evaluate our proposed method, which encompass a wide array of scenarios with diverse multi-label characteristics, and the details of which are summarized in Table \ref{tab:table1}. For each dataset, we employ an 80\%/10\%/10\% train/validation/test split for running the comparative methods. Specifically, NUS-WIDE\footnote{NUS-WIDE: \url{https://lms.comp.nus.edu.sg/wp-content/uploads/2019/research/nuswide/NUS-WIDE.html}} provides 161,789 training images and 107,859 validation images. We divide NUS-WIDE evaluation set into 35,593 validation examples and 72,266 test examples, and then use the original training set for training.

\begin{table}[t]
    \centering
    \renewcommand\arraystretch{1.3}
    
    \begin{tabular}{l|c c c c c}
    \hline
         Dataset    &\# Instance   &\#Feature   &\# Label &  L/I\\
    \hline\hline
         NUS-WIDE  &269,648   &225  &81 &1.9\\
         yeast  &2,417   &103  &14 &4.2 \\
         scene  &2,407   &294  &6 &1.1 \\
        CAL500  &502   &68  &174 &26.0\\ 
        Image  &2,000   &294  &5 &1.2\\
        corel5k  &5,000   &499  &374  &3.5\\
        corel16k-s1  &13,766   &500  &153  &2.9\\

        iaprtc12   &19,627 &1,000  &291  &5.7\\

        espgame   &20,768  &1,000  &268  &4.7\\

        mirflickr   &24,581   &1,000  &38  &4.7\\

        tmc2007   &28,596 &981  &22  &2.2\\
    \hline
    \end{tabular}
    \caption{The detailed characteristics for each dataset. L/I denotes Labels/images, i.e., the average number of positives per image.}
    \label{tab:table1}
\end{table}

\begin{figure}[t]
  \centering
   \includegraphics[width=1.12\linewidth]{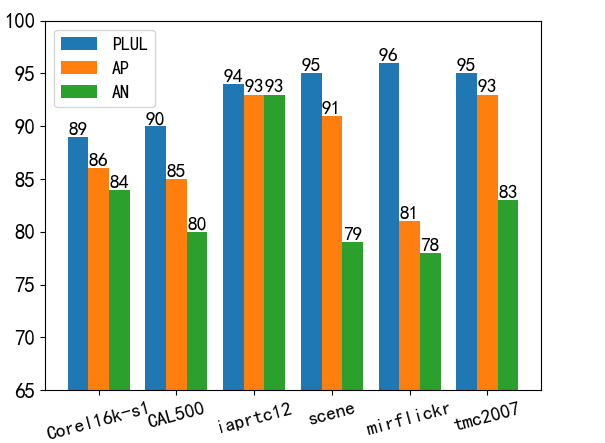}

   \caption{The classification accuracy(\%) of PLU labels which trained by using PLUL loss, AN loss and AP loss, respectively.}
   \label{fig:fig2}
\end{figure}

\textbf{Baseline and comparing methods.} We adopt AN loss and AP loss as the baselines, which treats the labels of PLU as negative or positive directly. For comparing the methods, we found that there are no previous work on solving multi-label with privacy-label. To evaluate the effectiveness of our method, we compare it with the following state-of-the-art weakly supervised multi-label classification approaches in the respect of their performance on concealed labels. GDF, which is the method of MLCLL, learns the classifier by nnotating negative labels \cite{MLCLL23}. Smile, which is the method of SPML, trains the model using single positive label via constructing an adjacency matrix to recover the latent soft label \cite{17}. Due to memory constraints, it is challenging to construct the matrix when the dataset is very large. Therefore, NUS-WIDE is not experiment on Smile. CLML, which is the method of MLML, introduces a label correction mechanism to identify missing labels, it can accurately make images close to their true positive and false negative images and away from their true negative images \cite{MLML21}. In addition to those, we include the fully supervised approach as a comparative to comprehensively evaluate the performance of our method. Furthermore, we computed the average accuracy of label classification in PLU, displayed in Table \ref{fig:fig2}.

\textbf{Setup.} For fairness, we employed the single-layer linear model as the prediction model for all the compared approaches. We train models using SGD optimizer with weight decay candidates $\left\{ 1e-3, 1e-4\right\}$ and the batch size is selected from $\left\{8, 16,32,64,256\right\}$. The learning rate is consider in range of $\left\{ 1e-1,1e-2,1e-3\right\}$, where the learning rate is multiplied by 0.1 at $\left\{40,60,100\right\}$ epochs \cite{MLCLL23}. The models are trained for 120 epochs on each datasets. All methods are evaluated performance using five widely recognized multi-label metrics: \textit{Ranking loss}, \textit{Hamming loss}, \textit{One-error}, \textit{Coverag}, and \textit{Average precision}\cite{MLLzhongsu24}. As for the fully supervised method, we use the same linear model as the other methods and standard BCE loss for training. 
To address the MLLPL problem, we individually crafted label sets for each comparative method: for Smile, we generate single positive label set in a dataset that hides sensitive labels; for GDF, we created a complementary label set by removing sensitive labels; for CLML, we directly used sensitive labels as missing labels to train. Due to variations in the number of classes for each comparing method, we set up two PLUs to train for all methods, i.e., each instance is associated with two PLUs. NUS-WIDE is trained with five PLUs.

\begin{table}[t]
\setstretch{1.25}
\begin{tabular}{p{1.7cm}|c c c c c}
\Xhline{2pt}
\multirow{3}{*}{CAL500}                           & \multicolumn{1}{c}{PLU}    & 2         &8         & 15                  & 30                   \\
\Xcline{2-6}{0.1pt}

                            & \multicolumn{1}{c}{BCE}& \multicolumn{1}{c}{47.27}  & \multicolumn{1}{c}{46.77} & \multicolumn{1}{c}{43.82} & \multicolumn{1}{c}{41.28}      \\
                           & ours                    &\textbf{ 48.44}          & \textbf{47.96}           &    \textbf{45.66 }                & \textbf{43.04}                              \\
\Xhline{1.5pt}
\multirow{3}{*}{yeast}          & \multicolumn{1}{c}{PLU}    & 2                  & 3                  & 4                  &5 \\
\Xcline{2-6}{0.1pt}
     & BCE                     & 68.40                    & 63.37                    & 50.39                    & 44.54                       \\
                           & ours                    & \textbf{76.70}                 &\textbf{ 73.58}                 & \textbf{63.67}                 & \textbf{58.06}                  \\
\Xhline{1.5pt}
 \multirow{3}{*}{tmc2007}           & \multicolumn{1}{c}{PLU}    & 2                  & 4                  & 6                  &8 \\
\Xcline{2-6}{0.1pt}
   & BCE                     & 68.23               & 60.09                 & 44.21                 & 32.18                    \\
                           & ours                    & \textbf{86.90 }                & \textbf{76.54}                 & \textbf{56.88}                 & \textbf{50.31}                    \\
\Xhline{1.5pt}
\multirow{3}{*}{mirflickr}            & \multicolumn{1}{c}{PLU}    & 2                  & 8                  & 10                  &15 \\
\Xcline{2-6}{0.1pt}
 & BCE                     & 60.70               & 45.25                     & 42.80                    & 20.51                  \\
                           & ours                    & \textbf{64.92}                 & \textbf{63.40 }                & \textbf{63.04}                & \textbf{62.84}                   \\
\Xhline{2pt}
\end{tabular}\caption{Accuracy comparison between BCE loss and our method with different PLU seting. The best results are in bold.}
\label{tab:table2}
\end{table}

\begin{table*}[t]
\centering
% \resizebox{1\textwidth}{
\begin{tabular}{c | c c c | c c c | >{\columncolor{gray}} c}

\hline
 Methods &GDF   &Smile   &CLML  &PLUL   &AN   &AP   &\cellcolor{gray!10}Fully-Supervised\\
\hline
  \multicolumn{8}{c}{Average Precision↑}\\
\hline
NUS-WIDE &20.77±1.52&- &50.23±0.31 &\textbf{51.30±0.11} &44.23±0.10 &51.07±0.16    &\cellcolor{gray!10}52.40±0.08\\
yeast   &48.89±0.32 &71.64±0.34 &76.62±0.14 &\textbf{76.70±0.10 }&73.44±0.83&73.63±0.42   &\cellcolor{gray!10}77.79±0.13\\
scene    &45.34±1.15 &66.22±3.84 &84.53±0.46 &\textbf{85.29±0.72} &66.46±0.67 &84.06±0.72   &\cellcolor{gray!10}84.48±0.31\\  
iaprtc12  &20.07±1.81  &23.82±0.42  &30.31±0.46  &\textbf{44.37±0.32}  &44.25±0.15  &43.90±0.20   &\cellcolor{gray!10}39.90±0.23\\
espgame  &15.73±1.48  &23.74±0.32  &21.04±0.47 &\textbf{27.97±0.12}  &26.43±0.41  &27.17±0.23    &\cellcolor{gray!10}27.78±0.17\\
mirflickr  &50.49±1.26  &58.35±0.45  &52.31±1.77  &\textbf{64.92±0.26}  &59.92±0.11  &60.01±0.19   &\cellcolor{gray!10}65.66±0.34\\
tmc2007  &33.18±0.45  &79.75±0.37  &78.30±0.24  &\textbf{86.90±0.21}  &81.63±0.47  &86.03±0.36   &\cellcolor{gray!10}78.89±0.73\\                                              
CAL500  &30.01±1.28  &32.22±0.92  & \textbf{48.44±0.17}&47.84±0.17  &46.97±0.30  &46.82±0.25  &\cellcolor{gray!10}47.66±0.36\\
Image  &61.11±0.84  &58.72±5.18  &63.69±0.23 &\textbf{65.00±0.45}  &52.92±0.55  &53.29±0.61   &\cellcolor{gray!10}72.58±0.31\\
corel5k  &20.31±0.72  &27.87±0.35  &25.13±0.19  &\textbf{29.91±0.24}  &21.74±0.15  &21.96±0.34   &\cellcolor{gray!10}31.39±0.24\\
corel16k-s1  &27.62±1.04  &31.25±0.31  &\textbf{32.10±0.35}  &31.66±0.16  &31.07±0.17  &31.26±0.24   &\cellcolor{gray!10}34.85±0.10\\

%mediamill &0.3996±0.012  &Error  &Error  &Error    &Error \\
\hline

 \multicolumn{8}{c}{Hamming loss↓}\\
\hline
NUS-WIDE &7.99±1.01&- &3.61±0.12 &\textbf{3.07±0.14} &3.17±0.21 &3.21±0.23  &\cellcolor{gray!10}2.73±0.14 \\
yeast  &37.65±0.95 &37.37±1.12 &21.65±0.16  &\textbf{20.66±0.8} &22.82±0.87 &21.31±0.81  &\cellcolor{gray!10}20.77±0.15\\
scene  &38.83±1.14 &18.00±3.67 &10.51±0.38 &\textbf{10.28±1.16} &17.37±0.84 &11.94±1.13   &\cellcolor{gray!10}11.00±0.12\\
iaprtc12  &3.70±1.83  &2.55±0.85  &2.11±0.23 &\textbf{1.94±0.74}  &2.00±0.64  &2.01±0.62  &\cellcolor{gray!10}2.09±0.71 \\
espgame  &2.48±0.36  &2.36±0.42  &2.53±0.12  &\textbf{1.99±0.11}  &2.22±0.14  &2.11±0.16  &\cellcolor{gray!10}1.98±0.18\\
mirflickr  &13.40±0.44  &11.97±0.78  &12.97±0.33 &\textbf{10.79±0.25}  &12.48±0.41  &12.28±0.38  &\cellcolor{gray!10}10.67±0.26 \\
tmc2007  &19.03±0.83  &6.56±1.84  &6.83±0.27 &\textbf{5.58±0.25}  &6.20±0.76  &5.88±0.64  &\cellcolor{gray!10}6.68±0.34 \\
CAL500 &29.43±1.25  &25.81±1.54  &\textbf{14.17±0.17} &15.29±0.10  &16.61±0.12  &16.66±0.21    &\cellcolor{gray!10}15.40±0.14\\
Image  &38.00±1.02  &48.96±4.15  &34.50±0.14 &\textbf{32.51±0.83}  &53.50±3.21  &52.50±2.64  &\cellcolor{gray!10}22.30±0.20 \\
corel5k &4.55±0.74  &\textbf{3.22±0.22}  &3.48±0.17 &3.37±0.13  &4.22±0.24  &4.21±0.37    &\cellcolor{gray!10}2.87±0.14\\
corel16k-s1  &3.85±0.95  &2.47±1.47  &\textbf{2.34±0.12} &2.38±0.11  &2.88±0.32  &2.57±0.24  &\cellcolor{gray!10}2.25±0.10\\

%mediamill &0.0660±0.007  &Error  &Error  &Error  &Error  &Error  &Error\\
\hline

\multicolumn{8}{c}{One-error↓}\\
\hline
NUS-WIDE &92.19±0.42&- &62.67±0.45 &\textbf{59.98±0.11} &61.65±0.15 &61.70±0.13  &\cellcolor{gray!10}60.81±0.23 \\
yeast  &86.31±1.63 &23.47±0.61 &20.80±0.23 &\textbf{22.73±1.54} &25.31±2.52 &23.61±2.57  &\cellcolor{gray!10}21.49±0.13 \\
scene  &85.00±1.75 &52.16±3.67 &26.14±0.42 &\textbf{24.75±1.35} &44.63±1.97 &26.08±3.36  &\cellcolor{gray!10}28.43±0.66 \\
iaprtc12  &68.42±2.37  &66.71±1.70  &58.02±0.74 &\textbf{39.53±1.32 } &44.19±1.34  &41.86±0.86  &\cellcolor{gray!10}46.36±0.21 \\
espgame  &96.00±2.02  &69.62±1.22  &81.61±0.56 &\textbf{60.70±2.14}  &62.07±1.01  &62.07±1.36   &\cellcolor{gray!10}63.02±2.30\\
mirflickr  &45.45±1.01  &35.53±1.14  &45.28±3.04  &\textbf{30.13±0.62}  &38.15±0.86  &33.33±1.67 &\cellcolor{gray!10}29.24±0.24 \\
tmc2007  &93.75±3.22  &20.88±0.47  &24.55±0.44 &\textbf{8.33±0.16}  &18.18±0.64  &8.33±0.13   &\cellcolor{gray!10}24.83±0.31\\
CAL500   &20.00±2.17  &21.96±15.45  &\textbf{13.07±0.95} &13.73±0.01  &13.87±0.27  &14.35±0.12  &\cellcolor{gray!10}13.73±0.18 \\
Image   &65.00±2.53  &64.70±7.47  &\textbf{48.33±0.59 } &52.78±1.12  &75.02±3.41  &70.29±2.34 &\cellcolor{gray!10}44.00±1.20 \\
corel5k  &96.88±1.76  &64.84±1.02  &74.27±0.16  &\textbf{63.56±0.73}  &85.62±1.66  &84.62±1.03 &\cellcolor{gray!10}82.60±0.69 \\
corel16k-s1  &88.89±2.04  &72.73±1.01  &\textbf{68.55±1.29} &70.27±0.23  &73.07±0.31  &72.25±0.44  &\cellcolor{gray!10}63.86±0.15\\

%mediamill &0.8333±0.063  &Error  &Error  &Error  &Error  &Error  &Error\\
\hline
\end{tabular}
%}
\caption{Experiment results (mean±std\%) for Average precision, Hamming loss and One-error on test data, where each instance is associated with one PLU. The optimal performance for each dataset is highlighted in bold. ↑/↓ denotes for higher performance for higher values and lower performance for lower values, respectively. The performance of fully supervised is presented in italics.}
\label{tab:table3}
\end{table*}

\begin{table*}[t]
\centering
\begin{tabular}{c | c c c | c c  c | >{\columncolor{gray}}c}
\hline
 Methods &GDF   &Smile   &CLML  &PLUL   &AN   &AP  &\cellcolor{gray!10}Fully-Supervised\\
\hline
\multicolumn{8}{c}{Coverage↓}\\
\hline
NUS-WIDE &19.44±0.18  &-  &14.35±0.14  &\textbf{13.93±0.16}  &27.96±0.22  &15.94±0.16 &\cellcolor{gray!10}12.98±0.33 \\
yeast   &55.06±0.39  &59.73±1.02  &48.67±0.29  &\textbf{48.44±0.74}  &49.12±1.28  &49.78±1.02  &\cellcolor{gray!10}43.73±0.17\\
scene    &29.86±0.98  &23.11±4.02  &9.13±0.58  &\textbf{8.40±0.92}  &29.55±1.34  &9.33±1.15 &\cellcolor{gray!10}8.37±0.18 \\         
iaprtc12  &72.66±0.82  &50.32±0.77  &35.61±0.74  &\textbf{25.09±0.26}  &26.52±0.43  &26.45±0.22  &\cellcolor{gray!10}26.55±0.14\\
espgame  &40.22±1.12  &53.27±1.65  &45.67±4.04  &\textbf{37.32±0.16}  &42.20±0.88  &41.20±0.64 &\cellcolor{gray!10}37.94±0.57 \\
mirflickr  &44.98±2.35  &41.47±1.02  &43.61±0.86  &\textbf{32.02±0.24}  &40.15±0.79  &38.40±1.76  &\cellcolor{gray!10}31.40±0.87\\
tmc2007  &17.90±4.14  &15.09±0.45  &15.44±0.35  &13.26±0.22  &13.86±4.05  &\textbf{13.06±0.21}  &\cellcolor{gray!10}15.17±0.36\\
CAL500  &82.61±1.24  &94.93±1.11  &\textbf{74.60±0.15}  &75.88±0.01  & 76.33±0.03 &76.58±0.08 &\cellcolor{gray!10}76.07±0.03 \\
Image  &31.00±1.57  &35.79±4.24  &\textbf{30.43±0.46 } &\textbf{29.98±0.32}  &44.00±1.02  &42.50±0.61  &\cellcolor{gray!10}20.10±0.43\\
corel5k  &26.80±1.30  &53.32±1.36  &46.32±0.17  &\textbf{25.24±0.44}  &66.48±2.23  &46.49±1.71&\cellcolor{gray!10}61.39±1.29 \\
corel16k-s1  &37.89±1.05  &32.63±0.52  &\textbf{29.51±0.26}  &32.09±0.15  &32.40±0.17  &32.33±0.13  &\cellcolor{gray!10}27.34±0.14\\

%mediamill &0.1287±0.080  &Error  &Error  &Error  &Error  &Error  &Error\\
\hline

\multicolumn{8}{c}{Rank Loss↓}\\
\hline
NUS-WIDE &15.14±0.19  &-  &9.92±0.14  &\textbf{9.17±0.13}  &16.35±0.14  &10.80±0.22  &\cellcolor{gray!10}8.97±0.30\\
yeast   &30.78±0.13  &22.05±0.47  &17.10±0.14  &\textbf{16.85±1.05}  &19.24±1.31  &19.61±0.85 &\cellcolor{gray!10}14.90±0.14 \\
scene   &44.61±1.23  &3.47±3.67         &8.68±1.66  &\textbf{8.36±0.52}  &33.38±1.06  &9.37±1.24 &\cellcolor{gray!10}8.52±0.27\\
iaprtc12  &24.07±3.15  &18.03±0.25  &12.63±1.33  &\textbf{8.27±0.30}  &8.39±0.33  &8.80±0.13 &\cellcolor{gray!10}8.91±0.12 \\
espgame  &18.04±4.23  &21.64±0.79  &19.29±1.40  &\textbf{15.58±0.17 } &17.26±0.77  &15.90±0.21  &\cellcolor{gray!10}15.68±0.41\\
mirflickr  &15.49±1.60  &15.01±0.65  &17.64±0.92  &\textbf{11.52±0.22}  &16.89±0.71  &14.4±0.29 &\cellcolor{gray!10}11.19±0.61\\
tmc2007  &14.96±2.27  &6.72±0.27  &7.09±0.75  &\textbf{5.01±0.37}  &6.49±0.47  &5.76±0.34  &\cellcolor{gray!10}6.40±0.22\\
CAL500  &31.97±1.15  &40.02±0.85  &\textbf{18.85±0.12}  &18.97±0.17  &19.13±0.27  &19.29±0.31  &\cellcolor{gray!10}18.19±0.11\\
Image  &31.25±0.84  &35.99±5.16  &39.15±0.47  &\textbf{29.17±0.35}  &46.88±0.45  &47.08±92  &\cellcolor{gray!10}21.00±0.52\\
corel5k  &26.25±0.67  &21.99±0.74  &\textbf{20.21±0.17}  &20.34±0.92  &25.64±0.66  &24.48±0.43 &\cellcolor{gray!10}19.89±0.11 \\
corel16k-s1  &11.34±1.04  &17.67±0.22  &\textbf{14.97±0.25}  &17.76±0.10  &18.04±0.24  &17.82±0.41 &\cellcolor{gray!10}13.87±0.13\\
%mediamill &0.0476±0.021  &Error  &Error  &Error  &Error  &Error  &Error\\
\hline
\end{tabular}
\caption{Experiment results (mean±std\%) for Coverage and Rank Loss on test data, where each instance is associated with one PLU. The optimal performance for each dataset is highlighted in bold. ↑/↓ denotes for higher performance for higher values and lower performance for lower values, respectively. The performance of fully supervised is presented in italics.}
\label{tab:table4}
\end{table*}\

\subsection{Experimental results} In Table \ref{tab:table2}, 
 we present a comprehensive comparison between our proposed method and the comparative approaches, employing three standard metrics: \textit{Average Precision}, \textit{Hamming Loss}, and \textit{One-error}. Regarding average precision, our results demonstrate competitive performance across all datasets. Specifically,  PLUL achieves an average accuracy of $55.62\%$ across the entire set of datasets, demonstrating a significant superiority over Smile by $12.75\%$. This notable advantage is attributed to the fact that our method is more suitable for dealing with private labeled data than SPML approaches. Moreover, PLUL exhibits comparable or even superior performance to CLML, especially, our \textit{Average Precision} on iaprtc12 surpasses CLML by $14.06\%$, underscoring the robustness and effectiveness of our approach. Those results prove our method can work well on MLLPL with missing labels problem.

Through Tables \ref{tab:table3} and \ref{tab:table4}, we can see that  PLUL consistently outperforms GDF across all evaluation metrics, which indicates that the MLCLL approach is insufficient in effectively addressing the challenges posed by the MLLPL problem. This is because the GDF implementation is based on the assumption that all labels can be selected as complementary labels, whereas privacy-labels are hidden in the MLLPL setting. That leads to the MLCLL method  could not capture the information of complementary labels for privacy-labels in MLLPL.

It is noteworthy that most datasets have accuracies approaching, or even exceeding fully supervised accuracy, e.g., scene, iaprtc12, tmc200. This remarkable performance can be attributed to the high classification accuracy of PLU labels in these datasets, reaching approximately $95\%$ as illustrated in Figure \ref{fig:fig2}. which indicates that a substantial number of concealed labels have been successfully restored. Thereby, our method achieves comparable accuracy even compared with fully supervised learning. Those results validate the effectiveness of our method to solve the MLLPL problem.

Furthermore, we find AN exhibits unfavorable performance in almost all cases, which demonstrates that simply treating the unobserved labels as negative labels cannot effectively solve MLLPL problems. Especially when the data contains very few classes, the detrimental impact of false negative labels introduced by AN becomes more pronounced. For instance, as shown in Table \ref{tab:table1}, datasets such as scenes and Image, each with only 6 and 5 classes, respectively,  show suboptimal accuracy when trained using the AN loss.
Specifically, the accuracy are only $66.46\%$ and $52.92\%$, representing a substantial $18.83\%$ and $12.08\%$ lower accuracy compared to PLUL, respectively. These results indicate that the effectiveness of AN is limited when a significant proportion of false negatives is introduced. This further underscores the superior performance of our proposed PLUL method in addressing the challenges posed by MLLPL problems.

To better validate the effectiveness of our algorithm, we separately list the classification accuracy of PLU labels in Figure \ref{fig:fig2}. It can be seen that our method is superior to AN loss and AP loss in all datasets. The average classification accuracy in PLU labels trained by PLUL loss reaches $93\%$, especially scene, mirflickr and tmc2007 even exceed $95\%$. By comparison we find that the results of AN are not satisfactory, which is consistent with our conclusionThe experimental results demonstrate the effectiveness of our proposed method in identifying hidden labels.
 %the positive contribution to model training分析RF

\subsection{Ablation studies}
In this section, we conducts the ablation study of PLUL loss ($\mathcal{L}_{\text{PLUL}}$), comparing it with BCE loss ($\mathcal{L}$) under different PLU settings. The results are shown in Table \ref{tab:table2}. When PLU is set to 2, the achieved performance using $\mathcal{L}_{\text{PLUL}}$ is $48.44\%$, $76.70\%$, $86.90\%$, and $64.92\%$ on the four datasets. These results are all superior to the performance achieved by utilizing $\mathcal{L}$, which is $47.27\%$, $68.40\%$, $68.23\%$, and $60.70\%$. As we increase PLU to 30, 5, 8, and 15 on the four datasets, our method exhibits a more pronounced advantage compared to the BCE approach, surpassing BCE by $1.76\%$, $13.52\%$, $18.13\%$, and $42.33\%$, respectively. This discloses $\mathcal{L}_{\text{PLUL}}$ helps the model recover concealed positive labels, thereby enhancing the training process. 

To validate the effectiveness of our method in concealing different numbers of labels, we set four different PLU for each dataset.  It can be observed that, with PUL increasing from 2 to 5 in yeast dataset, the performance using BCE loss witnesses a decline of $24\%$. In contrast, our method experiences only an $18\%$ decrease. Especially on the mirflickr, when PLU increases from 2 to 15, the performance with BCE loss droppe by $39.19\%$, while our method only experience a $2.18\%$ decrease. This indicates that our approach demonstrates robustness and stability when dealing with varying numbers of privacy labels. Even in scenarios with a high proportion of concealed labels, our method consistently maintains excellent accuracy. The experimental results convincingly validate the positive contribution of our method to addressing the challenges posed by the MLLPL problem. This robustness in performance underscores the adaptability of our approach in concealing privacy labels across different settings and datasets.

\section{Conclusion}
The paper studies the problem of multi-label learning with privacy-labels, which aims to conceal sensitive labels. In order to solve MLLPL problems, we design the PLU to conceal privacy-labels by combining a privacy-label with a randomly sampled non-privacy label. We propose PLUL loss function based on PLU labels, which is beneficial for identifying the concealed labels. The extensive experimental results on multiple benchmark datasets validate the effectiveness of our proposed method.

This work tentatively solves the problem of multi-label learning with privacy protection. We thank that this work has important real-life implications. In future work, we plan to incorporate regularization methods to further enhance the performance of MLLPL and continue exploring more efficient approaches to privacy label settings.

{
    \small
    \bibliographystyle{ieeenat_fullname}
    \bibliography{main}
}
% \bibliographystyle{unsrtnat}
% \bibliography{main}


\end{document}